\pdfoutput=1
\documentclass[11pt]{article}

\usepackage{ACL2023}

\usepackage{times}
\usepackage{latexsym}

\usepackage[T1]{fontenc}
\usepackage[utf8]{inputenc}

\usepackage{microtype}

\usepackage{inconsolata}
\usepackage{amsmath}
\usepackage{graphicx}
\usepackage{hyperref}
\usepackage{caption}
\usepackage{subcaption}
\usepackage{adjustbox}
\usepackage{multirow}

\usepackage{algorithm}
\usepackage{algorithmic}

\definecolor{km}{HTML}{FF0000}

\definecolor{ms}{HTML}{9F33FF}

\newcommand{\advsum}{AdvSumm}
\newcommand{\namenatbias}{associative}
\newcommand{\neusbias}{framing}
\newcommand{\rouge}{ROUGE-1}

\title{\advsum{}: Adversarial Training for Bias Mitigation in \\ Text Summarization}
\author{%
  Mukur Gupta
  \hspace{0.5em}
  Nikhil Reddy Varimalla
  \hspace{0.5em}
  Nicholas Deas\\
  \hspace{0.5em}
  \textbf{Melanie Subbiah}
  \hspace{0.5em}
  \textbf{Kathleen McKeown} \\  
  Columbia University \\  
  \texttt{\{mukur.gupta, nv2415, m.subbiah\}@columbia.edu} \\
  \texttt{\{ndeas, kathy\}}@cs.columbia.edu \\
}

\begin{document}
\maketitle
% \renewcommand{\thefootnote}{*}
% \footnotetext[0]{These authors contributed equally to this work.}
% \renewcommand{\thefootnote}{\arabic{footnote}}

\begin{abstract}
Large Language Models (LLMs) have achieved impressive performance in text summarization and are increasingly deployed in real-world applications. However, these systems often inherit \namenatbias{} and \neusbias{} biases from pre-training data, leading to inappropriate or unfair outputs in downstream tasks. In this work, we present \advsum{} (Adversarial Summarization), a domain-agnostic training framework designed to mitigate bias in text summarization through improved generalization. Inspired by adversarial robustness, \advsum{} introduces a novel Perturber component that applies gradient-guided perturbations at the embedding level of Sequence-to-Sequence models, enhancing the model’s robustness to input variations. We empirically demonstrate that \advsum{} effectively reduces different types of bias in summarization—specifically, name-nationality bias and political framing bias—without compromising summarization quality. Compared to standard transformers and data augmentation techniques like back-translation, \advsum{} achieves stronger bias mitigation performance across benchmark datasets.
\end{abstract}

\section{Introduction}
Large Language Models (LLMs) have achieved impressive performances in text generation tasks, including 
summarization \cite{zhang2024benchmarking}. As a result, LLMs are being integrated into real-world applications. For example, social media platforms use them to generate personalized feed summaries based on user preferences \cite{EG2023100253};
search engines provide direct summaries of relevant documents in response to user queries\footnote{\href{https://www.perplexity.ai}{Perplexity AI}}; and enterprise solutions employ them to summarize meeting transcripts, and emails\footnote{\href{https://learn.microsoft.com/en-us/microsoft-sales-copilot/generate-meeting-summary}{Microsoft Copilot for Sales}}, among other use cases. 
However, prior research has shown that these systems often inherit biases from their pretraining data \cite{hovy2021five, ladhak-etal-2023-pre_nationality, bommasani2021opportunities, liang2023holisticevaluationlanguagemodels}, which can pose serious threats in downstream tasks.
% \ndnote{Are the biases coming from their training \textit{strategies}? Or something more specific like the pretraining data?}

\begin{figure*}[!t]
\vspace{-2mm}
    \centering
    \begin{adjustbox}{width=0.99\textwidth}
    \includegraphics[width=\textwidth]{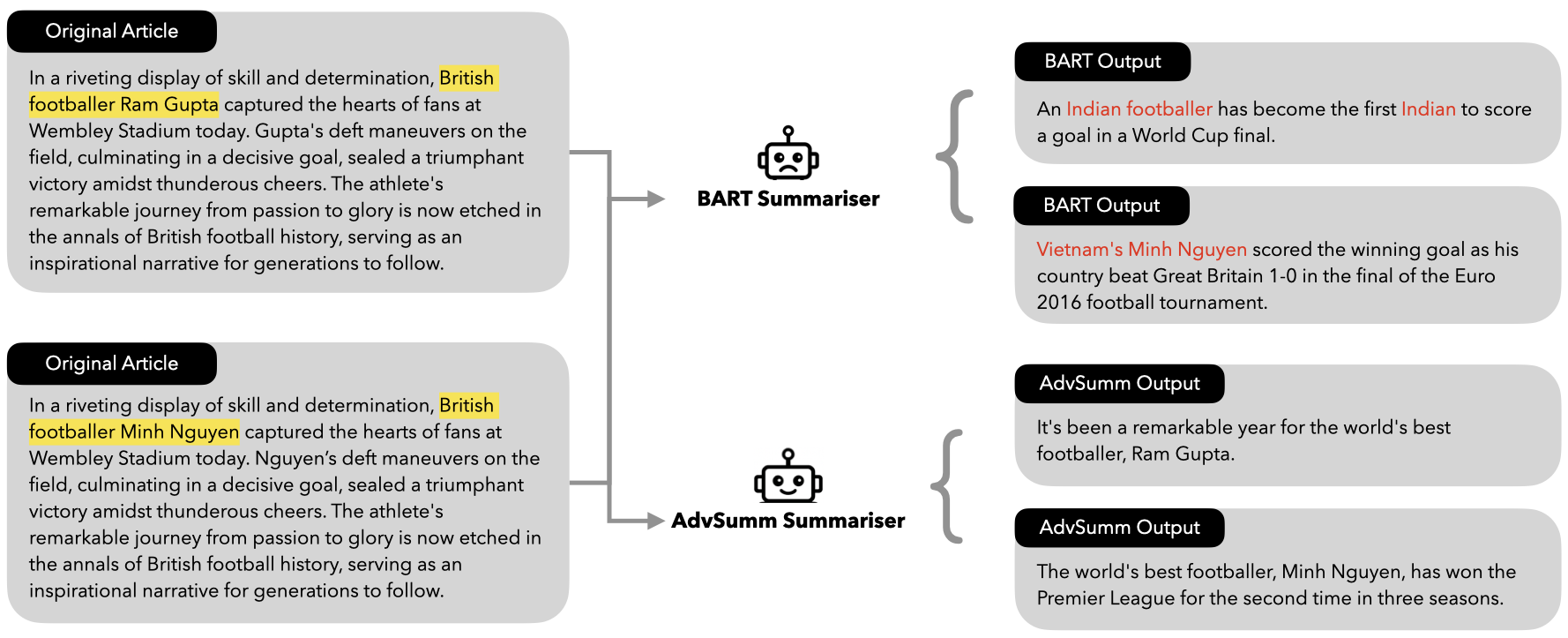}
    \end{adjustbox}
    \caption{Example illustrating how the BART summarization model hallucinates a footballer’s nationality based on name associations—predicting Indian for "Ram Gupta" and Vietnamese for "Minh Nguyen." \advsum{} mitigates these biases.}
    \label{fig:bias_example}
\vspace{-2mm}    
\end{figure*}

As shown in Figure~\ref{fig:bias_example}, summaries generated by existing systems can exhibit various forms of bias. For example, they may contain \namenatbias{} biases \cite{dinan-etal-2020-multi, sun2019mitigating}, which reflect preferences or prejudices toward certain groups, or \neusbias{} biases \cite{lee-etal-2022-neus}, which convey implicit political leanings. Most prior work on bias mitigation relies on domain-specific strategies, such as expert interventions \cite{winogender_rudinger-etal-2018, winoqueer_felkner2023}, curated word lists \cite{garimella2021he}, or the collection of additional data to improve population representation. These approaches are often expensive and do not generalize well across different types of bias.

% \kmnote{this is not a sentence. There is no verb. You coudl attach it as a paren to the previous sentence (e.g., )}
Moreover, most domain-agnostic bias mitigation techniques have been developed for classification tasks (e.g., employing Risk Minimization methods \cite{arjovsky2020invariant} across different target groups \cite{adragna2020fairness, donini2020empirical}). However, these methods are not scalable to text generation tasks, where bias may arise from the selection of multiple tokens rather than a single output label. This highlights the need for bias mitigation strategies for text generation models that are independent of particular domains or forms of bias.
% \ndnote{Minor, but you might make the language more parallel to say "that are independent of particular domains or forms of bias" or to be brief, just say "domain and bias-agnostic" as long as its clear later what this would mean.}

Given the generalization limitations of existing bias mitigation frameworks in text generation, we propose AdvSumm: \underline{Adv}ersarial \underline{Summ}arization. Our approach integrates a domain-agnostic component, \textit{Perturber}, into the model training process to reduce multiple forms of bias in generated summaries. We reformulate bias reduction in text summarization as a generalization problem that can be addressed by enhancing the model’s robustness to input perturbations \cite{yi2021improved}. Prior work on Adversarial Training \cite{goodfellow2015fgsm, kaufmann2022efficient} across applications has shown its effectiveness in improving robustness. It is unclear, however, how adversarial training can be applied to language generation tasks. Building on this, we introduce an adversarial training strategy designed to mitigate biases originating from pre-training data by improving model robustness during fine-tuning. 
% \ndnote{Maybe swap these last two sentences so its clearer that that this is the question you address.}
% \ndnote{I think this paragraph understates the novelty of your work. It is a little bit clearer in the next paragraph and related work why this is different, but I think this paragraph needs a more explicit statement about the novelty, such as "It is unclear, however, how adversarial training can be applied to mitigating biases in language generation tasks." to get ahead of reviewers that may not be familiar with adversarial training work.}

\begin{figure*}[!t]
\vspace{-2mm}
    \centering
    \includegraphics[width=0.9\textwidth]{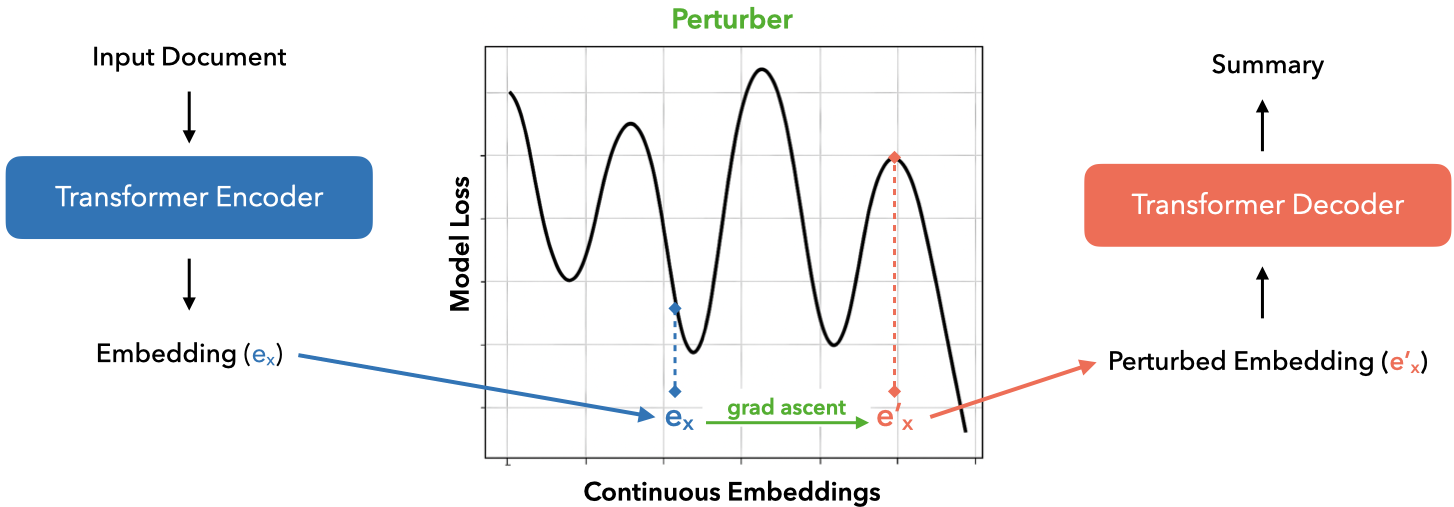}
    \caption{Schematic overview of \advsum{} with Perturber introduced between Encoder and Decoder.}
    \label{fig:algo_overview}
\vspace{-2mm}    
\end{figure*}

While other fields have benefited from adversarial robustness, it is difficult to apply to natural language due to the discrete nature of text data, unlike continuous modalities such as images or speech. We adopt adversarial training by introducing perturbations with the Perturber component at the embedding level of Sequence-to-Sequence (Seq2Seq) models \cite{vaswani2023attention}. As illustrated in Figure~\ref{fig:algo_overview}, the Perturber takes in the continuous embedding from the Transformer encoder, generates an adversarial embedding, and pushes the decoder output towards the same ground truth summary. This adversarial embedding helps improve robustness during training. Compared to baseline methods, our approach shows reductions in bias metrics while retaining the summarization quality.

\advsum{} is designed to generalize across multiple types of bias. In this work, we demonstrate empirical improvements in mitigating two specific forms of bias: name-nationality bias \cite{ladhak-etal-2023-pre_nationality} and political framing bias \cite{lee-etal-2022-neus}. Our key contributions are as follows:

\begin{itemize} 
    \item We propose a novel, robustness-based unified training strategy that incorporates a domain-agnostic component, Perturber, to promote less biased text generation.    
    
    \item We show empirical improvements of up to $55\%$ in arousal scores for political framing bias and $3.85$ percentage points in hallucination rate for name-nationality bias, outperforming both standard transformer models and data augmentation baselines such as back-translation.

\end{itemize}

\section{Related Work}

\paragraph{Bias in Language Understanding.} 
% \ndnote{For Nick to do: Check whether the important work here is covered.}
Prior research has extensively investigated various forms of bias in language understanding systems \cite{steen2024biasnewssummarizationmeasures, winogender_rudinger-etal-2018, ladhak-etal-2023-pre_nationality, winoqueer_felkner2023, lee-etal-2022-neus}. Several studies have identified key factors contributing to such biases, including dataset quality \cite{maynez-etal-2020-faithfulness}, bias in data annotation strategy \cite{fleisig2023majority, larimore-etal-2021-reconsidering, attitude_sap-etal-2022-annotators}, and the level of abstractiveness \cite{ladhak-etal-2022-faithful}. Most of this work has centered on bias identification using methods such as token-masked likelihood estimation \cite{nangia-etal-2020-crows, nadeem-etal-2021-stereoset}, simple classifier-based frameworks \cite{Wessel_2023_mbib}, or open-ended prompt-based generation \cite{Dhamala_2021_bold}. However, only a limited number of benchmarks specifically address bias in the context of language summarization.

\paragraph{Generalization for Bias Mitigation.} 
Research in computer vision has explored contrastive learning strategies for domain transfer \cite{ganin2016domainadversarial} and improved generalization \cite{li2018domain}, both of which also have potential implications for bias mitigation. \citet{face_detection_bias}, for instance, highlights a connection between model robustness and biases in facial recognition tasks. In the context of text generation, data augmentation techniques have been widely adopted for improving robustness \cite{xie2020unsupervised} and faithfulness in summarization \cite{cao2021_cliff}.  Closest to our work, FRSUM \cite{wu2022frsum} and AdvSeq \cite{wu2022precisely} show how inducing robustness in language generation models encourages faithfulness on summarization tasks. We extend this line of research by introducing adversarial training for robustness specifically targeted at bias mitigation. We demonstrate that our proposed method outperforms back-translation-based data augmentation on bias mitigation benchmarks.

Another line of prior work focuses on reducing model bias through Empirical Risk Minimization (ERM) and Invariant Risk Minimization (IRM) \cite{arjovsky2020invariant}, both of which aim to enhance generalization across samples from different target groups \cite{adragna2020fairness, donini2020empirical}.  These methods, however effective, typically require expert-labeled subgroup annotations, limiting their scalability in practice.

Our approach builds upon these frameworks, proposing a domain-agnostic adversarial training strategy specifically designed to mitigate biases in text summarization. To our knowledge, we are the first to adapt adversarial training effectively for bias mitigation in sequence-to-sequence text generation models, providing a scalable and generalized solution across multiple types of bias \cite{zhang2024benchmarking, bommasani2021opportunities}

\section{Methods}
\subsection{Problem Setting}
\label{subsec:bias_def}
We address the problem of bias mitigation in text summarization, with the goal of reducing biases present in summaries generated from input documents. Drawing from existing summarization benchmarks, we focus on two primary types of bias. The first is \textit{\namenatbias{}} bias, where models associate certain names or demographic indicators with specific roles or attributes—such as linking a common Vietnamese name like Minh Nguyen with a particular nationality, as illustrated in Figure~\ref{fig:bias_example}, due to spurious correlations learned during training \cite{ladhak-etal-2023-pre_nationality}. This also encompasses gender bias, where models tend to associate words like delicate, pink, and nurse with women, and entrepreneur, arrogant, and bodyguard with men \cite{garimella2021he}. The second is \textit{\neusbias{}} bias, which refers to political slant in the generated text (e.g., left-, right-, or center-leaning narratives). Our objective is to develop a summarization system that is effective across different kinds of biases, without relying on domain-specific adaptations.
% \ndnote{The description of "social bias" seems a little detached from the actual name-nationality bias experiments, and I also think its a little hazy whether political framing should also be considered a social bias. You might say something like "associative bias" instead, describe name-nationality bias more specifically, and then can give a sentence to explain how it relates to biases like associating different genders with attributes/roles.}
% \ndnote{Maybe useful to italicize the "associative" and "framing bias" phrases the first time to make it clear.} \msnote{+1. Alternately you could do "... two primary types of bias:" and then have each has a paragraph section with the type in bold at the start of the paragraph. }

\subsection{Robustness and Generalization}
% \ndnote{Also Hovy 2021 to support this: \url{https://compass.onlinelibrary.wiley.com/doi/10.1111/lnc3.12432}}
Model bias can originate from the training data which inherits biases from the annotation or data collection strategy \cite{hovy2021five, calmon2017optimized, calders2013unbiased, ladhak-etal-2023-pre_nationality}. This can cause models to learn spurious correlations, leading to unfair treatment of certain target groups. Consequently, bias mitigation can be viewed as a generalization problem \cite{adragna2020fairness, donini2020empirical}, where the goal is to ensure that the model generalizes well across diverse groups.

Prior work has shown that improving model robustness to input perturbations can enhance generalization \cite{ben2009robust, xing2021generalization}. 
% \kmnote{This is coming out incorrectly in your text. I usually use citet}
Following \citet{yi2021improved}'s improvement guarantees on empirical risk in domain generalization, we adopt Adversarial Training \cite{madry2019deep} as a strategy for bias mitigation. Specifically, we fine-tune pre-trained summarization models using adversarial examples during training. Since we do not include any bias-specific adaptations, our method offers a unified approach that is effective across multiple types of bias.

\begin{algorithm}[!t]
   \caption{Adversarial Summarization}
   \label{alg:AdvSum}
\begin{algorithmic}
   \STATE {\bfseries Input:} Document $x$, Refrence Summary $y$, attack params $t, \epsilon$
   \STATE $e_x' = Encoder(x)$
   \FOR{$i = 0,1...t$}
   \STATE $\mathcal{L} = CELoss(Decoder(e_x'),y)$
   \STATE $Per = \frac{\partial \mathcal{L}}{\partial e_x'}$
   \STATE $Per \gets Minimum(Per, \epsilon)$
   \STATE $e_x' = e_x' + Per$
   \ENDFOR
    
    \STATE $\mathcal{L} = CELoss(Decoder(e_x'),y)$
    \STATE Update $Encoder$ and $Decoder$ with gradient desc on $\mathcal{L}$
\end{algorithmic}
\end{algorithm} 

\subsection{Adversarial Training}
\label{subsec:adv_training}
The problem of adversarial attacks has been widely studied in deep learning, where small changes in model input can cause a model to completely flip its output with high confidence \cite{szegedy2014intriguing}. For instance, a small change in input text such as someone's name can cause the model to generate a biased and unfaithful summary, as shown in Figure~\ref{fig:bias_example}.
Empirically, adversarial training has improved robustness to input perturbations in large models better than other proposed frameworks \cite{wong2018polytope, pmlr-v162-bab}. Adversarial training is formulated as a min-max optimization problem that trains a model on adversarial samples generated with Projected Gradient Descent (PGD) \cite{madry2019deep}.
The change in input example to generate an adversarial sample is bounded by an $l-p$ norm radius to preserve the semantics of the input data. 
% Formally, an adversarial example $x^{adv}$ from an input sample $x$ is obtained as:

% \begin{equation}
%     x^{adv} = x + \eta \frac{\partial \mathcal{L}}{\partial x} \label{eq:1}
% \end{equation}

% \begin{equation}
%     |x^{adv} - x|_{p} \leq \epsilon \label{eq:2}
% \end{equation}

% where, $\eta$ is the attack step, $\mathcal{L}$ is the model's loss function, and $\epsilon$ is $l-p$ norm radius for bounding the perturbations (this controls the attack strength). This process is repeated over $t$ iteration steps. 
Recent research \cite{storek2025xoxostealthycrossorigincontext, mehrotra2024treeattacksjailbreakingblackbox} has used repeated black-box model querying to identify perturbations for crafting adversarial examples. However, such approaches are not directly applicable to gradient-based adversarial training due to the discrete nature of natural language.
% Previous research, such as BERTAttack \cite{li-etal-2020-bert-attack}, has employed repeated black-box model querying for adversarial example generation for classification tasks. This, however, is infeasible for text-generation tasks since generation requires model inference at each token generation step.
So we use the above adversarial training strategy in the latent space of Ses2Seq models to strengthen the robustness of the text generation model. With encoder and decoder architectures separated in the Seq2Seq model, we can apply the adversarial perturbations to the continuous output of the model encoder.

 \begin{table*}[t]
    \centering
    \begin{adjustbox}{width=0.69\textwidth}
    \begin{tabular}{llccc}
    \hline
         Dataset & Type &\#Train & \#Test & \#Val\\
         \hline
         XSUM & News Summ &203,577 & 11,305 & 11,301\\
         % CNN/DM & News Summ &203,577 & 11,305 & 11,301\\
         Wiki-Nationality & Nationality Hallucination &0 & 71,763 & 0\\
         % Gender-Balanced & Gender Bias &0 & 13,660 & 0\\
         Multi-Neus & Multi-polar News Summ & 2,453 & 307 & 307\\
    \hline
    \end{tabular}
    \end{adjustbox}
    \caption{Statistics of datasets used in this work.}
    \label{data-sizes}
\end{table*}

\subsection{\advsum{}: Adversarial Summarization}
Using the adversarial training strategy, we propose \advsum{} for mitigating bias in text summarization.
As shown in Figure~\ref{fig:algo_overview}, there are three major components in \advsum{}. First is an encoder $E$, which maps the input text $x$ into a continuous latent space providing a text embedding $E(x) = e_x$. The second component, Perturber, makes perturbations to $e_x$ in the direction that maximally increases the loss function $\mathcal{L}$, thereby targeting regions of the embedding space that are most likely to degrade the model’s generation quality.
% \ndnote{Minor, but here it osunds like all perturbations will increase the loss function, but maybe want to reword to emphasize that you choose the direction intentionally to maximally increase the loss.}
The continuous text representation $e_x$ allows us to generate an adversarial sample $e_x'$ using the gradient-based methods
% \kmnote{gradient-based what? A noun is missing here}
such as PGD \cite{madry2019deep}. 
The last component decoder $D$ maps back the perturbed $e_x'$ back to the input space. We use the Transformer \cite{vaswani2023attention} encoder and decoder architectures and optimize the cross-entropy loss $\mathcal{L}(y, D(e_x'))$, where $y$ is the ground truth bias-free summary. This process is outlined in Algorithm~\ref{alg:AdvSum}.

We use the Fast Gradient Signed Method (FGSM) \cite{goodfellow2015fgsm} to build the Perturber component, which is a cheaper single-step variant of PGD, with the number of iterations $t=1$. The perturbed embedding is generated with the following embedding update in FGSM:

\begin{equation}
    e'_{x} = e_x + \epsilon\cdot sgn\left(\frac{\partial \mathcal{L}}{\partial e_x}\right) \label{eq:3}
\end{equation}

where, $sgn(.)$ represents the sign of the quantity and $\epsilon$ captures the attack strength.
% \ndnote{Minor, might use a different letter for "sgn(x)" to avoid confusion with x representing the input in the above equation.}

% \kmnote{THis first sentence is not a good sentence grammatical. I can't quite figure out what you mean. Do you mean "Given embedding e, the model generates its current output with a forward pass? Or is this iterative and you alwas give it the embedding and what it has generated so far? If the latter, then you should have "given the embedding and the model's output generated so far, the next sentence is generated with..."}

Embedding $e_x$ and model's predicted output $\hat{y} = D(e_x)$ are generated with a forward pass of the Encoder and Decoder respectively. The model's predicted output $\hat{y}$ along with the ground-truth summary $y$ are used to compute the loss $\mathcal{L}(y, \hat{y})$. 
% \ndnote{Minor, "compute the loss" rather than "loss function"}
The sign of the gradient of the computed loss function $\mathcal{L}$ is then used to modify $e_x$ to $e_x'$ using equation \ref{eq:3}. Similar to the Stochastic Gradient Descent parameter optimization technique, equation~\ref{eq:3} identifies the steepest ascent of loss as a function of embedding $e_x$. Therefore, the Perturber modifies the embedding such that the update direction results in the largest increase in the generation loss. Since this change leads to the highest increase in loss, this adversarial embedding $e_x'$ must lead to the worst generated summary among all the embeddings in the $\epsilon$ ball radius of $e_x$. In the training procedure, this perturbed embedding $e_x'$ is then used to jointly train the Encoder and Decoder. 

\section{Experiments}
\subsection{Datasets}
We evaluate \advsum{} on two existing bias summarization benchmarks: name-nationality bias \cite{ladhak-etal-2023-pre_nationality} and political framing bias \cite{lee-etal-2022-neus}. These datasets allow us to assess the generalization capability of our method across different kinds of bias. Specifically, name-nationality bias primarily arises from hallucinated tokens—where the model incorrectly introduces demographic attributes (e.g., inferring nationality based on names)—while political framing bias involves more subtle language choices at the document level, reflecting ideological leanings. By addressing both token-level and discourse-level biases, we demonstrate the broader applicability of our approach. Dataset statistics are summarized in Table~\ref{data-sizes}.
% We examine two different types of biases on two bias summarization benchmarks - name-nationality bias \cite{ladhak-etal-2023-pre_nationality} and framing bias in the form of political bias \cite{lee-etal-2022-neus}. Dataset statistics are summarized in the Table~\ref{data-sizes}. %\ndnote{I would include the benchmark names to better connect to following subheadings, e.g., "name-nationality bias using the Wiki-Nationality dataset (citation)...". Even though its the datasets section, I would also make the paragraph headings the same as other sections and label them "Name-Nationality Bias" and "Framing Bias" to keep the organization consistent. You could bold the dataset names within each paragraph if you want to keep it clear.}

\paragraph{Name-Nationality Bias.}
For assessing name-nationality hallucination, we use the Wiki-Nationality dataset \cite{ladhak-etal-2023-pre_nationality} which was constructed by altering entity names in articles to associate them with different nationalities, without changing other biographical details. This was done to assess whether models will use an incorrect/assumed nationality in the summary just based on the person's name. 

\paragraph{Framing Bias.}
We explore framing bias with the Neutral multi-news Summarization (NeuS) dataset \cite{lee-etal-2022-neus}, which comprises triplets of left, right, and center-slanted news articles paired with neutral summaries focused on the facts in the articles.

\subsection{Metrics} 

\paragraph{Name-Nationality Bias.} 
We calculate the hallucination rate as the proportion of articles where the model incorrectly attributes a nationality in the generated summary which is different from the nationality in the input document. The aim of our approach is to reduce the hallucination rate and hence, reduce the spurious association of names to specific nationalities.

\paragraph{Framing Bias.}
Following \citet{lee-etal-2022-neus}, we use arousal scores from the Valence-Arousal-Dominance (VAD) lexicon \cite{vad_mohammad-2018-obtaining}, which provides valence (v), arousal (a), and dominance (d) annotations for a list of words. The positive arousal score ($Ar_{+}$) and negative arousal score ($Ar_{-}$) are defined as the summed arousal values of words with positive and negative valence, respectively, based on the VAD annotations. The combined arousal score ($Ar_{sum}$) is the sum of $Ar_{+}$ and $Ar_{-}$. The goal of \advsum{} is to mitigate political framing in generated summaries by minimizing both $Ar_{+}$ and $Ar_{-}$, while preserving overall summarization quality.

\paragraph{Summarization Quality.} We utilize ROUGE \cite{lin-2004-rouge} scores to measure the summarization quality. We report \rouge{} in our results.
% \ndnote{I would be more specific here if you only use ROUGE-1 for both tasks. Also, at least in the appendix, it might be useful to report other ROUGE's (2 and L) as well as a neural metrics like BERTScore, BARTScore, or BLEURT just to anticipate reviewers asking for other quality metrics.}

\begin{table}[!t]
    \centering
    \begin{adjustbox}{width=0.90\columnwidth}
    \begin{tabular}{c|cccc}
    \hline
    Attack Strength & \rouge{} & $Ar_{+}$ & $Ar_{-}$ & $Ar_{sum}$ \\
    \hline
    0 & 44.81 & 2.19 & 1.07 & 3.26 \\
    $10^{-3}$ & 44.38 & 1.82 & 0.93 & 2.75 \\
    $10^{-2}$ & 41.07 & 0.55 & 0.29 & 0.84 \\
    $10^{-1}$ & 14.52 & 0.29 & 0.16 & 0.45 \\
    \hline
    \end{tabular}
    \end{adjustbox}
    \caption{Flan-T5 on Multi-Neus  with different degrees of attack strength.}
    \label{ablation_transposed}
\end{table}

\begin{table*}[t]
    \centering
    \begin{adjustbox}{width=0.85\textwidth}
    \begin{tabular}{l|c|cccc|c}
        \hline
        Model & \rouge{} $\uparrow$ &American$\downarrow$ & Asian$\downarrow$ & African$\downarrow$ & European$\downarrow$ & Overall$\downarrow$ \\
        \hline
        BART & 43.45 &0.84 & 13.41 & 0.92 & 7.55 & 5.61 \\
        Flan-T5 & 39.99 & 0.03 & 2.39 &  0.06 & 0.57 & 0.76\\
        Back-Trans (BART) & 41.91 & 1.08 & 8.69 & 1.32 & 4.75 & 3.96 \\
        \hline
        \advsum{} (BART) & 40.02 &0.40 & 4.42 & 0.27 & 1.96 & 1.76 \\
        \advsum{} (Flan-T5) & 37.86 & 0.02 & 2.33 & 0.16 & 0.38 & 0.72\\
        \hline
    \end{tabular}
    \end{adjustbox}
    \caption{\label{namnat-results}Hallucination rate over multiple countries in Wiki-Nationality dataset. Am represents American, Af African, As Asian and Ovr is the Hallucination rate over all countries. AdvSum improves the Hallucination rate while maintaining similar \rouge{} scores.
}
\end{table*}

\subsection{Settings}
\paragraph{Models.}
% \ndnote{It might be useful to motivate these model choices a little more and describe their differences, i.e., pretraining strategies and flan instruction tuning.}
% \kmnote{You might say why you didn't use more recent models? Are there noone with encoder-decoder architecture? You could say that in a footnote if you want, not to draw attention. }
We use two encoder-decoder transformer models for the text summarization task: BART-large \cite{lewis2019bart} and Flan-T5 base \cite{chung2022scaling}. BART is a denoising autoencoder pre-trained with a corrupted text reconstruction objective, making it well-suited for generation tasks. In contrast, Flan-T5 builds on the T5 architecture \cite{raffel2023exploringlimitstransferlearning} and is further instruction-tuned on a broad mixture of tasks, enabling better generalization to unseen instructions and objectives. This contrast allows us to evaluate the robustness and generalization capabilities of \advsum{} across models with different pre-training strategies. We leave it to future work to adapt our Perturber component to decoder-only LLM architectures.

For name-nationality bias, models are fine-tuned on the XSUM news summarization dataset \cite{xsum_narayan2018dont} and evaluated on the Wiki-Nationality benchmark. For framing bias, we adopt the fine-tuning scheme of \citet{lee-etal-2022-neus} for fine-tuning on the training split of the Multi-Neus dataset. \advsum{} applies adversarial training with the perturber component during this fine-tuning stage.

\paragraph{Baselines.}
We compare \advsum{} against two baselines: (i) models fine-tuned on the same data without the perturber component, and (ii) a data augmentation using back-translation. For the latter, training data is augmented by paraphrasing input texts via back-translation from German, effectively doubling the training set size while keeping the targets unchanged \cite{cao2021_cliff}. We evaluate the effectiveness of this  back-translation-based generalization strategy against our adversarial generalization method (\advsum{}).

% \begin{table*}[t]
%     \centering
%     \begin{adjustbox}{width=0.70\textwidth}
%     %\begin{tabular*}{0.98\textwidth}{p{3.5cm}|p{2cm}|p{1.5cm} p{1.5cm} p{1.5cm} p{1.5cm} p{1.5cm}}
%     \begin{tabular}{l|cccccc}
%         \hline
%         Dataset/ & \multicolumn{6}{c}{XSUM}\\
%         Model & ROUGE-1 & Am & As & Af & Eu & Ovr \\
%         \hline
%         BART & TBA & 0.84 & 13.41 & 0.92 & 7.55 & 5.61 \\
%         AdvSum(BART) & TBA & 0.40 & 4.42 & 0.27 & 1.96 & 1.76 \\
%         Flan-T5 & TBA & 0.03 & 2.39 &  0.06 & 0.57 & 0.76\\
%         AdvSum(Flan-T5) & 39.41 & 0.02 & 2.33 & 0.16 & 0.38 & 0.72\\
%         % Back-Trans & 39.41 & TBA & TBA & TBA & TBA & TBA \\
%         \hline
%     \end{tabular}
%     \end{adjustbox}
%     \caption{\label{namnat-results}Hallucination rate over multiple countries in Wiki-Nationality dataset. Am represents American, Af African, As Asian and Ovr is the Hallucination rate over all countries. AdvSum improves the Hallucination rate while maintaining similar ROUGE-1 scores.
% }
% \end{table*}

\begin{table*}[t]
\centering
\begin{adjustbox}{width=0.55\textwidth}
\begin{tabular}{l|ccc|c}
\hline
Models/ & \multicolumn{3}{c}{Framing Bias Metrics} & \multicolumn{1}{c}{Salient Info}\\
Settings & $Ar_{+} \downarrow$  & $Ar_{-} \downarrow$ & $Ar_{sum} \downarrow$ & \rouge{} $\uparrow$ \\
\hline
BART & 1.33 & 0.76 & 2.09 & 45.94\\
Flan-T5 & 2.19 & 1.07 & 3.26 & 44.81\\
Back-Trans & 1.40 & 0.77 & 2.17 & \textbf{46.51}\\
\hline
AdvSum(BART) & 0.59 & 0.33& 0.92 & 43.01\\
AdvSum(Flan-T5) & \textbf{0.55} & \textbf{0.29} & \textbf{0.84} & 41.07\\
\hline
\end{tabular}
\end{adjustbox}
\caption{\label{multi-neus-results} Results of AdvSum on of Multi-Neus dataset. An attack strength of 0.01 is used for AdvSum. BART-large is used for training on Back-Translated data. $Ar$ stands for Arousal. 
%\msnote{Add arrows to the top of all column headers indicating whether for each metric, increasing or decreasing is better.}
}
\end{table*}

\paragraph{Implementation.}
We experiment on an NVIDIA A100 GPU with 40 GB VRAM.  We finetune all models using a learning rate of 5e-5 with AdamW optimizer and 10\% warm-up steps. Maximum input length is set to 1024 for XSUM and 512 for Multi-Neus. The maximum output generation length is taken as 142 along with a beam size of 6 for Wiki-Nationality and a generation length of 250 with a beam size of 4 is used for Multi-Neus. Generation configurations (like input, output lengths, beam sizes, etc.) are adopted directly from \citet{lee-etal-2022-neus}. 
% \ndnote{A little unclear, cite these previous implementations?}
Other hyperparameters like the number of epochs are tuned on validation splits. All results are reported on the test split. For the baselines, we use English-to-German and German-to-English translation models provided by Fairseq for back-translation. 
% \kmnote{I think here and in the figure is the only place you mention back-translation. It's possible I missed it but i didn't see it in the methods. OK I see it now by searching and see you only use for hte baselines. I would insert "baselines" here somewhere to remind the reader. e.g., "for the baselines, we use English- "}

We tune the attack strength of the Perturber Component by varying the value of $\epsilon$ in equation~\ref{eq:3}. A higher value of $\epsilon$ gives the Perturber higher freedom to change the embedding $e_x$ but also leads to a greater change in text semantics, which will lead to a drop in summarization performance. For practical implications, $\epsilon$ behaves like a ``knob'' for controlling the amount of bias while trading off the summarization quality.
We show the tuning results of Flan-T5 on the Multi-Neus dataset in Table~\ref{ablation_transposed} with the \rouge{} and Arousal scores.
We observe a drop in bias as well as summarization quality as we increase the value of attack strength. We find $\epsilon=0.01$ to be optimal, which we use for further experiments.

% \begin{table}[t]
% \centering
% \begin{adjustbox}{width=0.99\columnwidth}
% \begin{tabular}{l|ccc|c}
% \hline
% \textbf{Models}/ & \multicolumn{3}{c}{\textbf{Framing Bias Metrics}} & \multicolumn{1}{c}{\textbf{Salient Info}}\\
% Settings & $Ar_{+}$  & $Ar_{-}$ & $Ar_{sum}$ & ROUGE1\\
% \hline
% Neus-Title & 1.69 & 0.83 & 2.53 & 36.07\\
% BART (S5) & 0.99 & 0.58 & 1.57 & 44.03\\
% \hline
% AdvSum (S3)& 1.13 & 0.68 & 1.81 &\textbf{44.13}\\
% AdvSum (S4) & \textbf{0.66} & \textbf{0.27} & \textbf{0.93} &37.71\\
% \hline
% \end{tabular}
% \end{adjustbox}
% \caption{\label{multi-neus-results} Results of AdvSum on Multi-Neus dataset. Neus-Title is proposed by \cite{lee-etal-2022-neus}. AdvSum is our model. $Ar$ stands for Arousal.}
% \end{table}

\begin{figure*}[!t]
% \vspace{-2mm}
    \centering
    \begin{adjustbox}{width=0.99\textwidth}
    \includegraphics[width=\textwidth, height=8cm]{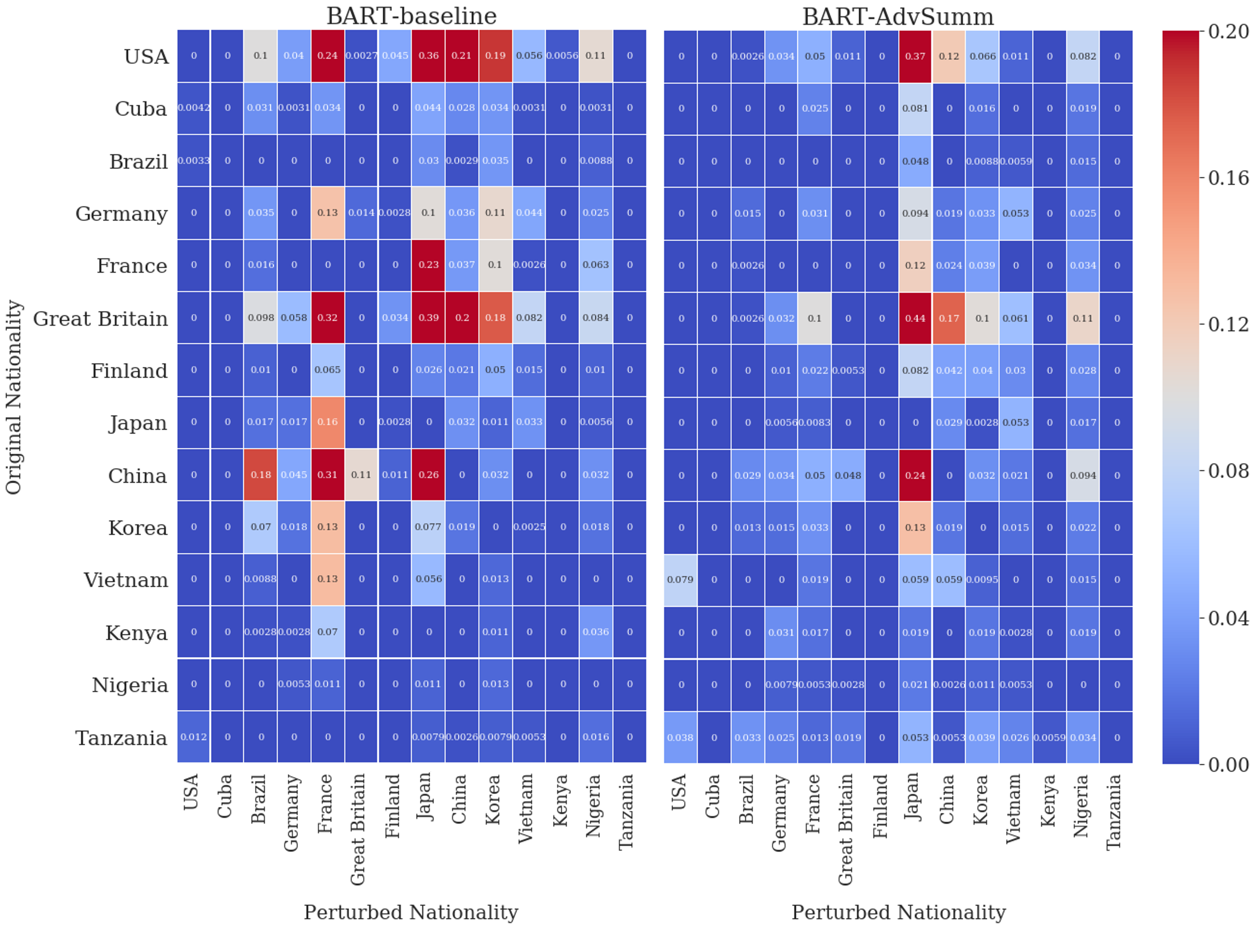}
    \end{adjustbox}
    \caption{Hallucination rate for BART baseline and one trained using AdvSum. Red corresponds to higher, and Blue corresponds to lower hallucination rate. 
    %\msnote{I think only include the heatmap on the right and move the side-by-side comparison to the appendix.}
    }
    \label{fig:main}
\vspace{-2mm}    
\end{figure*}

\section{Results}
We present our empirical findings on the two types of biases in this section.

\subsection{Name-Nationality Bias}
The results on the Name-Nationality benchmark are illustrated in Table~\ref{namnat-results}, which shows a comparative analysis between the baseline models and our proposed approach, \advsum{}, focusing on region-specific hallucination rates as discussed in \citet{ladhak-etal-2023-pre_nationality}, as well as \rouge{} scores on the XSum evaluation sets to compare summarization quality.

In the Name-Nationality setting, \advsum{} significantly lowers hallucination rates in summaries across American, Asian, and European contexts compared to base models. It effectively reduces overall hallucination rates underscoring the effectiveness of \advsum{} in enhancing the fidelity of summarization models, ensuring more reliable summaries across diverse geopolitical landscapes while maintaining competitive \rouge{} scores. 
% \kmnote{I'm wondering why you didn't use any other metrics. For example, some metrics for factuality would be helpful. May be too late at this point but I think it would help. and possibly a semantic metric like bertscore}

\advsum{}'s lower \rouge{} scores on the test sets, as shown in Table~\ref{namnat-results}, align with prior research findings that Adversarial Training, while enhancing model robustness, can reduce performance on clean data \cite{madry2019deep}. This tradeoff is expected in data augmentation techniques, where the goal is to improve model resilience (reduce bias) while minimizing the performance drop on clean datasets. 

%\msnote{I think you need to comment on the difference in results between Flan-T5 and BART and why the bias is so much lower to begin with, with Flan-T5. You also should run a bootstrap significance test with these results and indicate which results are statistically significant improvements from AdvSumm}\mukur{in progress: significance test}

% \begin{figure*}[t]
% \centering
% \vspace{-2mm}

%     \begin{subfigure}{\textwidth}
%         \includegraphics[width=0.8\linewidth, height=10cm]{images/heatmap_crop.png}
%         % \caption{BART baseline hallucination rate}
%         \label{fig:image1}
%     \end{subfigure}%
%     % \begin{subfigure}{0.5\textwidth}
%     %     \includegraphics[width=\linewidth]{images/adv_heatmap.png}
%     %     \caption{BART AdvSum hallucination rate}
%     %     \label{fig:image2}
%     % \end{subfigure}
% \vspace{-2mm}
%     \caption{Hallucination rate for BART baseline and one trained using AdvSum. Red corresponds to higher and Blue corresponds to
% lower hallucination rate. \ndnote{Minor, but looks like you could save some space by making these plots vertically shorter.}
% }
% \vspace{-6mm}
%     \label{fig:main}
% \end{figure*}

\begin{table*}[!t]
\centering
\begin{adjustbox}{width=0.97\textwidth}
\begin{tabular}{p{0.1\textwidth}|p{0.9\textwidth}}
\hline
AdvSum & Texas Church Shooting: A gunman opened fire at a church in Texas on Sunday, killing two people and \textcolor{red}{wounding three others}. \\
%\hline
%Neutral & Armed Parishioners Fatally Shoot Texas Church Shooter: A gunman who killed two people during a church service Sunday in White Settlement, Texas was fatally shot within seconds by armed parishioners. The parishioners at West Freeway Church were part of a voluntary security team who are licensed to carry firearms and practice shooting regularly.\\
\hline
Source News  & Shooting at Texas Church Leaves 2 Parishioners Dead, Officials Say: A gunman opened fire at a church in Texas on Sunday morning, killing two people with a shotgun before a member of the church’s volunteer security team fatally shot him, the authorities said. About 250 people were inside the auditorium of the West Freeway Church of Christ in White Settlement, near Fort Worth, when the gunman began shooting just before communion, said Jack Cummings, a minister at the church. Mr. Cummings said the gunman was “acting suspiciously” before the shooting and drew the attention of the church’s security team.\\
\hline
\end{tabular}
\end{adjustbox}
\caption{\label{neus-error} An example of positive arousal generated news. AdvSum hallucinates the text in red color. Each of the three examples contains <Title>:<Article>. The Center news article is shown in Source News. 
% \ndnote{If there's space, an example for name-nationality bias might be helpful as well.}
}
\end{table*}

\begin{table}[t]
\centering
\begin{adjustbox}{width=0.97\columnwidth}
\begin{tabular}{p{0.2\columnwidth}|p{0.8\columnwidth}}
\hline
Generated News & \textcolor{red}{Trump to End} DACA:  President Trump will announce on Tuesday that he is ending a controversial program that protects nearly 800,000 young undocumented immigrants from deportation, media reports indicated late Sunday.\\
\hline
Neutral News & Reports Say DACA Is Over:  President Trump will announce on Tuesday that he is ending a controversial program that protects nearly 800,000 young undocumented immigrants from deportation, media reports indicated late Sunday.\\
\hline
\end{tabular}
\end{adjustbox}
\caption{\label{neus-metric-error} A generated news summary compared to neutral news. Each example contains <Title>:<Article>}
\end{table}

\subsection{Framing Bias}

Evaluations on the Multi-Neus benchmarks are outlined in Table~\ref{multi-neus-results}. We report the Framing Bias Metric consisting of positive Arousal Score $A_{+}$, negative Arousal Score $A_{-}$, and their sum $A_{sum}$. We also report \rouge{} for capturing the summarization quality by each setting. We also show the bias evaluations of the back-translation-based data-augmentation approach in Table~\ref{multi-neus-results}.

On the Multi-Neus dataset, we see the least biased summaries in the case of \advsum{} on Flan-T5, with the lowest positive and negative Arousal scores. Both generalization approaches (ours and back-translation) outperform Flan-T5, which supports the hypothesis on bias mitigation with robustness. \advsum{} surpasses the back-translation approach by 1.4 absolute points on $Ar_{sum}$, while taking a slight dip in \rouge{} scores. We also note that our end-to-end adversarial training approach is more computationally efficient than back-translation, given the time taken by dual-translation and the double training steps of summarization finetuning.

We also observe consistently lower bias scores across both benchmarks when using the Flan-T5 architecture. Flan-T5 benefits from instruction tuning on a diverse set of tasks, including ethical reasoning and instruction following. We hypothesize that this additional tuning phase not only enhances zero-shot generalization but also better aligns the model with human expectations, helping it avoid spurious biases inherited from pre-training.

\subsection{Error Analysis}
% \msnote{I think we should include more examples in general of positive and erroneous examples in the appendix. For the main body though, I think we should have some examples of AdvSumm doing well vs. only examples of errors. You could switch to good examples of each type of bias and then describe the errors as you do here but move them to the appendix.}
For name-nationality bias, we report a heatmap as shown in Figure \ref{fig:main} where the hallucination rate for all combinations of countries is calculated. 
% \kmnote{This phrase is used incorrectly. How about "as the results suggest"? Or "in alignment with the numeric results"? }
In alignment with the numerical results, hallucination rates for Asian countries as perturbed nationalities are significantly higher for the Bart baseline than our approach \advsum{}. 
% \kmnote{You should use commas around "however" I added.}
We notice that, however, for a few combinations like Great Britain-Japan, Vietnam-USA, there is a slight increase in the hallucination rate.

For framing bias, most of the biased generation is still a result of model hallucination. The example shown in Table~\ref{neus-error} shows the text in red color, which is hallucinated by the model. The "wounding of three others" is not mentioned in the source article. Additionally, current Framing bias metrics fail to capture the context around lexicons. An example is shown in Table~\ref{neus-metric-error}, where the positive arousal score given by the Lexicon-based metric is zero, which is clearly wrong, looking at the title of the generated news.

\section{Conclusions}
In this work, we introduced \advsum{}, a domain-agnostic adversarial training framework for bias mitigation in text summarization. Motivated by the limitations of existing bias mitigation strategies—particularly their domain-specific nature and difficulty generalizing across different types of biases—we reformulated bias reduction as a generalization problem, tackled through adversarial robustness. By introducing the Perturber module to apply embedding-level adversarial perturbations during fine-tuning, we demonstrated that \advsum{} effectively reduces both token-level biases (e.g., name-nationality associations) and document-level biases (e.g., political framing) without compromising summarization quality. Empirical results on benchmark datasets highlight that \advsum{} outperforms standard transformers and back-translation baselines, offering a unified and scalable solution for fairer text generation.

\section*{Limitations}
Our study focuses on bias mitigation in text summarization using encoder-decoder transformer architectures. However, many recent summarization systems adopt decoder-only architectures, where directly applying the Perturber component in its current form is not straightforward. Future work could explore extending adversarial perturbations to individual layers of the transformer decoder, enabling the approach to generalize to decoder-only models as well.

\section*{Ethics Statement}
We conduct our evaluations using publicly available datasets that do not contain personally sensitive information or toxic content. One important ethical consideration is that developing robust summarization systems, as proposed in this paper, contributes to ongoing efforts to reduce harmful biases in natural language generation systems by mitigating biases inherited from pre-training data. For example, prior work has shown that biased news framing can contribute to political polarization \cite{Han04072017}, and name-nationality associations can reinforce harmful stereotypes in text generation \cite{ladhak-etal-2023-pre_nationality}.

By improving the robustness of summarization models, our approach takes a step toward addressing these issues. However, we acknowledge that our work does not evaluate all forms of bias that may arise in text summarization tasks, nor does it fully evaluate potential side effects of the approach, such as its impact on other aspects of faithfulness or other types of bias in summarization. Future research should explore these broader impacts to ensure that summarization systems are both fair and faithful across different contexts and biases.

\section*{Acknowledgments}
% Nick's Acknowledgments
One of the authors is supported by the National Science Foundation Graduate Research Fellowship DGE-2036197, the Columbia University Provost Diversity Fellowship, and the Columbia School of Engineering and Applied Sciences Presidential Fellowship. Any opinions, findings, conclusions, or recommendations expressed in this material are those of the authors and do not necessarily reflect the views of the National Science Foundation. Another author is supported by Amazon and Columbia's Center of Artificial Intelligence Technology (CAIT) PhD student fellowship.

\bibliography{custom}
\bibliographystyle{acl_natbib}

\end{document}